\pdfoutput=1

\documentclass[11pt]{article}

\usepackage{acl}

\usepackage{times}
\usepackage{latexsym}

\usepackage[T1]{fontenc}

\usepackage[utf8]{inputenc}

\usepackage{microtype}

\usepackage{bm}
\usepackage{amsmath}
\usepackage{graphicx}
\usepackage{amsfonts}
\DeclareMathOperator*{\E}{\mathbb{E}}

\usepackage{booktabs}

\title{Parallel Attention Forcing for Machine Translation}

\author{Qingyun Dou \and Mark Gales \\
  University of Cambridge \\
  \texttt{\{qd212,mjfg100\}@cam.ac.uk} \\}

\begin{document}
\maketitle
\begin{abstract}
Attention-based autoregressive models have achieved state-of-the-art performance in various sequence-to-sequence tasks, including Text-To-Speech (TTS) and Neural Machine Translation (NMT), but can be difficult to train. The standard training approach, teacher forcing, guides a model with the reference back-history. During inference, the generated back-history must be used. This mismatch limits the evaluation performance. Attention forcing has been introduced to address the mismatch, guiding the model with the generated back-history and reference attention. While successful in tasks with continuous outputs like TTS, attention forcing faces additional challenges in tasks with discrete outputs like NMT. This paper introduces the two extensions of attention forcing to tackle these challenges. (1) Scheduled attention forcing automatically turns attention forcing on and off, which is essential for tasks with discrete outputs. (2) Parallel attention forcing makes training parallel, and is applicable to Transformer-based models. The experiments show that the proposed approaches improve the performance of models based on RNNs and Transformers.
\end{abstract}

\section{Introduction}
Attention-based models are good at connecting sequences of different length, and have achieved state-of-the-art performance in various sequence-to-sequence (seq2seq) tasks \cite{vaswani2017attention, tay2020efficient}. Here the term performance refers to the overall quality of the output sequences, e.g. word error rate in Automatic Speech Recognition (ASR). On the other hand, these models can be difficult to train \cite{bengio2015scheduled}. From a probabilistic perspective, seq2seq models estimate the probability of the output sequence conditioned on the input sequence. To achieve more accurate estimation, the models are often autoregressive \cite{chen2018best}. The standard training approach, teacher forcing, guides a model with reference back-history during training. This makes the model unlikely to recover from its mistakes during inference, where the model operates in free running mode, and the reference output is replaced by the generated output. This problem is referred to as exposure bias \cite{ranzato2015sequence}.

Many approaches have been introduced to address exposure bias, and will be described in section \ref{sec:intro_inf_train}. Attention forcing is a simple and effective option \cite{dou2020attention}. The idea is to guide the model with the generated output history and reference attention. While successful in TTS, attention forcing faces additional challenges when it comes to tasks with discrete outputs \cite{dou2019attention}, and models such as Transformers \cite{vaswani2017attention}. To tackle these challenges, this paper introduces scheduled attention forcing in section \ref{sec:ch_af_sched}, and parallel attention forcing in section \ref{sec:chAF_paf}. The experiments in section \ref{sec:chNMT_exp} show that these approaches improve strong NMT models based on RNNs and Transformers.\footnote{
Links to the source code for the experiments will be available after the anonymous review. 
} 


\section{Attention-based sequence-to-sequence generation}

Sequence-to-sequence generation can be defined as the task of mapping an input sequence $\bm{x}_{1:L}$ to an output sequence $\bm{y}_{1:T}$ \cite{bengio2015scheduled}. 
From a probabilistic perspective, a model $\bm{\theta}$ estimates the distribution of $\bm{y}_{1:T}$ given $\bm{x}_{1:L}$, which can be factorized into token distributions: $p(\bm{y}_{1:T}|\bm{x}_{1:L}; \bm{\theta}) = \prod_{t=1}^{T} p(\bm{y}_{t} | \bm{y}_{1:t-1}, \bm{x}_{1:L}; \bm{\theta})$.

\subsection{Encoder-attention-decoder architecture}

\begin{figure}
\centering
\includegraphics[width=7.5cm]{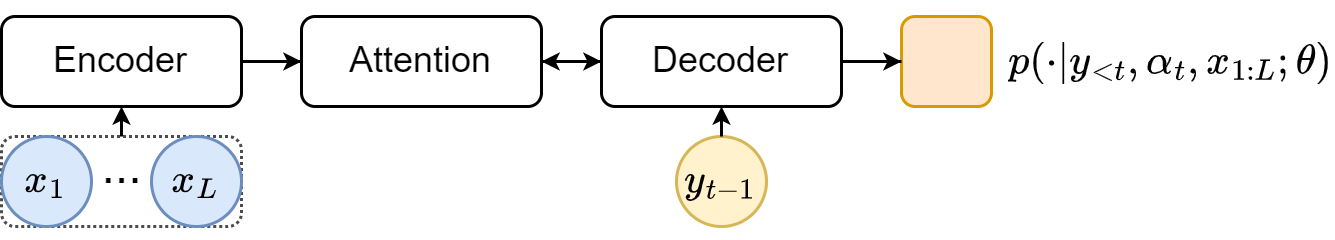}
\caption{An attention-based encoder-decoder model, in teacher forcing mode, at decoding step $t$; a circle depicts a token, and a rounded square a distribution.}\label{fig:EAD}
\end{figure}

Attention-based seq2seq models usually have the encoder-attention-decoder architecture \cite{lewis2020bart, tay2020efficient}. Figure \ref{fig:EAD} shows the architecture. The distribution of a token is conditioned on the back-history $\bm{y}_{1:t-1}$, input sequence $\bm{x}_{1:L}$ and an attention map $\bm{\alpha}_{1:T}$:
\begin{align}
p(\bm{y}_{t}|\bm{y}_{1:t-1}, \bm{x}_{1:L}; \bm{\theta}) &\approx p(\bm{y}_{t}|\bm{y}_{1:t-1}, \bm{\alpha}_{t}, \bm{x}_{1:L}; \bm{\theta}) \nonumber \\
&\approx p(\bm{y}_{t}|\bm{s}_{t}, \bm{c}_{t}; \bm{\theta}_{y}) \label{eq:py_general_p} 
\end{align}
where $\bm{\theta} = \{\bm{\theta}_{y}, \bm{\theta}_{s}, \bm{\theta}_{\alpha}, \bm{\theta}_{h}\}$; $\bm{s}_{t}$ is a state vector representing $\bm{y}_{1:t-1}$, and $\bm{c}_{t}$ is a context vector summarizing $\bm{x}_{1:L}$ for time step $t$. The discussions in this paper are agnostic to the form of attention. The following equations give a detailed example about how $\bm{\alpha}_{t}$, $\bm{s}_{t}$ and $\bm{c}_{t}$ can be computed:
\begin{gather}
\bm{h}_{1:L} = f(\bm{x}_{1:L}; \bm{\theta}_{h}) \label{eq:EAD_enc}\\
\bm{s}_{t} = f(\bm{y}_{1:t-1}; \bm{\theta}_{s}) \label{eq:EAD_s}\\
\textstyle \bm{\alpha}_{t} = f(\bm{s}_{t}, \bm{h}_{1:L}; \bm{\theta}_{\alpha}) \quad
\bm{c}_{t} = \sum_{l=1}^{L} \alpha_{t,l} \bm{h}_{l} \label{eq:EAD_att_2}\\
\hat{\bm{y}}_{t} \sim p(\cdot | \bm{s}_{t}, \bm{c}_{t}; \bm{\theta}_{y}) \label{eq:EAD_dec}
\end{gather}
The encoder maps $\bm{x}_{1:L}$ to $\bm{h}_{1:L}$, considering the entire input sequence; $\bm{s}_{t}$ summarizes $\bm{y}_{1:t-1}$, considering only the past. With $\bm{h}_{1:L}$ and $\bm{s}_{t}$, the attention mechanism computes $\bm{\alpha}_{t}$, and then $\bm{c}_{t}$. The decoder estimates a distribution based on $\bm{s}_{t}$ and $\bm{c}_{t}$, and optionally generates an output token $\hat{\bm{y}}_{t}$.

\subsection{Inference and training} \label{sec:intro_inf_train}

During inference, given an input $\bm{x}_{1:L}$, the output $\hat{\bm{y}}_{1:T}$ can be obtained from the distribution estimated by the model $\bm{\theta}$: $\hat{\bm{y}}_{1:T} = \underset{\bm{y}_{1:T}}{\mathrm{argmax}} \, p( \bm{y}_{1:T} | \bm{x}_{1:L}; \bm{\theta})$.
The exact search is often too expensive and approximated by greedy search for continuous output, or beam search for discrete output \cite{bengio2015scheduled}.

Conceptually, the model is trained to learn the natural distribution, e.g. through minimizing the KL-divergence between the natural distribution $p(\bm{y}_{1:T}|\bm{x}_{1:L})$ and the estimated distribution $p(\bm{y}_{1:T} | \bm{x}_{1:L}; \bm{\theta})$. In practice, this can be approximated by minimizing the Negative Log-Likelihood (NLL) over some training data $\{\bm{y}^{(n)}_{1:T}, \bm{x}^{(n)}_{1:L}\}_{1}^{N}$, sampled from the natural distribution:
\begin{align}
\mathcal{L}(\bm{\theta}) &= \E_{\scalebox{0.8}{$\bm{x}_{1:L}$}} \mathrm{KL} \big(p(\bm{y}_{1:T}|\bm{x}_{1:L}) || p(\bm{y}_{1:T}|\bm{x}_{1:L}; \bm{\theta}) \big) \nonumber \\
&\propto - \textstyle \sum_{n=1}^{N} \log p(\bm{y}^{(n)}_{1:T}|\bm{x}^{(n)}_{1:L}; \bm{\theta}) \label{eq:pb_train}
\end{align}
$\mathcal{L}(\bm{\theta})$ denotes the loss; $N$ denotes the size of the training dataset; $n$ denotes the data index. To simplify the notation, $n$ is omitted for the length of the sequences, although they also vary with $n$. In the following sections, the sum over the training set will also be omitted.

A key question here is how to compute the token distribution $p(\bm{y}_{t}|\bm{y}_{1:t-1}, \bm{x}_{1:L}; \bm{\theta})$. For the most standard training approach, teacher forcing, the token distribution is computed with the reference output history $\bm{y}_{1:t-1}$ at each time step $t$. The loss can be written as:
\begin{equation} \label{eq:loss_y_iclr}
    \mathcal{L}_{y}(\bm{\theta}) 
    = - \textstyle \sum_{t=1}^{T} \log p(\bm{y}_{t}| \bm{y}_{1:t-1}, \bm{x}_{1:L}; \bm{\theta})
\end{equation}
Despite its advantages such as parallel training \cite{dou2022improving}, teacher forcing suffers from exposure bias: during training, the model is guided by the reference output history; during inference, however, the model runs in free running mode, where the generated output history is used. This mismatch leads to errors that can accumulate along the inference process \cite{ranzato2015sequence}.

There are mainly two lines of research addressing exposure bias. Scheduled sampling \cite{bengio2015scheduled} and professor forcing \cite{lamb2016professor} are prominent examples along the first line. These approaches guide a model with both the reference and the generated output history, and the goal is to learn the data distribution. To facilitate convergence, they often depend on a heuristic schedule or an auxiliary classifier, which can be difficult to design and tune \cite{guo2019new}. The second line is a series of sequence-level training approaches, leveraging reinforcement learning \cite{ranzato2015sequence}, minimum risk training \cite{shen2016minimum} or generative adversarial training \cite{yu2017seqgan}. Theses approaches guide a model with the generated output history. During training, the model operates in free running mode, and the goal is not to generate the reference output, but to optimize a sequence-level loss. However, many tasks do not have well established sequence-level objective metrics. Examples include voice conversion, machine translation and text summarization \cite{tay2020efficient}. Both lines of research require sequentially generating output sequences during training. In recent years, Transformers \cite{vaswani2017attention} have been widely used, and parallel training has been essential. To efficiently generate output sequences from Transformer-based models, an approximation scheme \cite{duckworth2019parallel} has been proposed to parallelize scheduled sampling.

\section{Attention forcing 2}

\begin{figure}
\centering
\includegraphics[width=7.5cm]{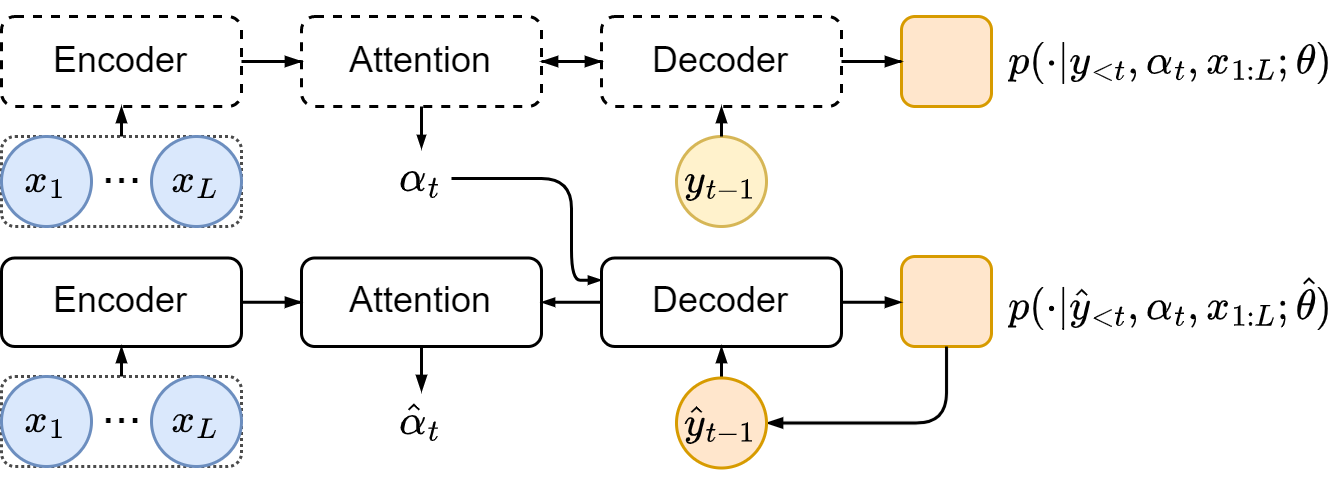}
\caption{Attention forcing; the solid blocks are the attention forcing model $\hat{\bm{\theta}}$; the dashed blocks the teacher forcing model $\bm{\theta}$. During inference, $\hat{\bm{\theta}}$ runs in free running mode without $\bm{\theta}$.} \label{fig:AF_coarse}
\end{figure}

This section will revisit the general framework of attention forcing, and then analyze its challenges and introduce extensions to tackle them.
The basic idea of attention forcing \cite{dou2020attention} is to train the model with the generated output and reference attention. The generated output helps addressing the exposure bias, and reference attention helps with convergence. Let $\bm{\theta}$ denote a standard attention-based model, trained with teacher forcing. Let $\hat{\bm{\theta}}$ denote a model with the same structure, but trained with attention forcing, and later used for inference. Figure \ref{fig:AF_coarse} illustrates attention forcing. In attention forcing mode, equation \ref{eq:py_general_p} becomes:
\begin{align} \label{eq:py_A}
p(\bm{y}_{t}|\bm{y}_{1:t-1}, \bm{x}_{1:L}; \hat{\bm{\theta}}) &\approx p(\bm{y}_{t}| \hat{\bm{y}}_{1:t-1}, \bm{\alpha}_{t}, \bm{x}_{1:L}; \hat{\bm{\theta}}) \nonumber \\
&\approx p(\bm{y}_{t} | \hat{\bm{s}}_{t}, \hat{\bm{c}}_{t}; \hat{\bm{\theta}}_{y})
\end{align}
$\hat{\bm{s}}_{t}$ and $\hat{\bm{c}}_{t}$ denote the state vector and context vector generated by $\hat{\bm{\theta}}$. Details of attention forcing are in the following equations:
\begin{gather}
\begin{matrix}
&\bm{h}_{1:L} = f(\bm{x}_{1:L}; \bm{\theta}_{h}) &\hat{\bm{h}}_{1:L} = f(\bm{x}_{1:L}; \hat{\bm{\theta}}_{h})
\end{matrix} \label{eq:AF_enc}\\
\begin{matrix}
&\bm{s}_{t} = f(\bm{y}_{1:t-1}; \bm{\theta}_{s}) &\hat{\bm{s}}_{t} = f(\hat{\bm{y}}_{1:t-1}; \hat{\bm{\theta}}_{s})
\end{matrix} \label{eq:AF_s}\\
\begin{matrix}
&\bm{\alpha}_{t} = \scalebox{0.8}{$f(\bm{s}_{t}, \bm{h}_{1:L}; \bm{\theta}_{\alpha})$} 
&\hat{\bm{\alpha}}_{t} = \scalebox{0.8}{$f(\hat{\bm{s}}_{t}, \hat{\bm{h}}_{1:L}; \hat{\bm{\theta}}_{\alpha})$} 
\end{matrix}  \label{eq:AF_att_1}\\
\hat{\bm{c}}_{t} =\textstyle \sum_{l=1}^{L} \alpha_{t,l} \hat{\bm{h}}_{l} \label{eq:AF_att_2}\\ 
\hat{\bm{y}}_{t} \sim p(\cdot | \hat{\bm{s}}_{t}, \hat{\bm{c}}_{t}; \hat{\bm{\theta}}_{y}) \label{eq:AF_dec} 
\end{gather}

The right side of the equations \ref{eq:AF_enc} to \ref{eq:AF_att_1}, as well as equations \ref{eq:AF_att_2} and \ref{eq:AF_dec}, show how the attention forcing model $\hat{\bm{\theta}}$ operates. The decoder state $\hat{\bm{s}}_{t}$ is computed with $\hat{\bm{y}}_{1:t-1}$. While an alignment $\hat{\bm{\alpha}}_{t}$ is generated by $\hat{\bm{\theta}}$, it is not used by the decoder, because the context $\hat{\bm{c}}_{t}$ is computed with the reference alignment $\bm{\alpha}_{t}$. One option of obtaining $\bm{\alpha}_{t}$ is shown by the left side of equations \ref{eq:AF_enc} to \ref{eq:AF_att_1}: to generate $\bm{\alpha}_{t}$ from a teacher forcing model $\bm{\theta}$. $\bm{\theta}$ is trained in teacher forcing mode, and generates $\bm{\alpha}_{t}$ in the same mode. Although the reference output is used to compute the reference attention, it is not directly fed into the model, hence the model cannot rely too much on the back-history.

At the inference stage, the attention forcing model operates in free running mode, and equation \ref{eq:AF_att_2} becomes $\hat{\bm{c}}_{t} = \textstyle \sum_{l=1}^{L} \hat{\alpha}_{t,l} \hat{\bm{h}}_{l}$. The decoder is guided by $\hat{\bm{\alpha}}_{t}$, instead of $\bm{\alpha}_{t}$.

During training, there are two objectives: to infer the reference output and to imitate the reference alignment. This can be formulated as:
\begin{align}
\mathcal{L}_{y,\alpha} (\hat{\bm{\theta}}) &= \mathcal{L}_{y} (\hat{\bm{\theta}}) + \gamma \mathcal{L}_{\alpha} (\hat{\bm{\theta}}) \label{eq:loss_joint_AF}\\
\mathcal{L}_{y}(\hat{\bm{\theta}}) &= -\scalebox{1.0}{$\sum_{t=1}^{T}$} \log p(\bm{y}{\scriptscriptstyle _{t}}|\hat{\bm{y}}{\scriptscriptstyle _{1:t-1}}, \bm{\alpha}{\scriptscriptstyle _{t}}, \bm{x}{\scriptscriptstyle _{1:L}}; \hat{\bm{\theta}}) \nonumber\\ 
\mathcal{L}_{\alpha} (\hat{\bm{\theta}}) &= \scalebox{1.0}{$\sum_{t=1}^{T}$} \mathrm{KL}(\bm{\alpha}_{t}||\hat{\bm{\alpha}}_{t}) \nonumber\\
&= \textstyle \sum_{t=1}^{T} \sum_{l=1}^{L} \alpha_{t,l} \log \frac{\alpha_{t,l}} {\hat{\alpha}_{t,l}} \nonumber
\end{align}
$\mathcal{L}_{y}$ and $\mathcal{L}_{\alpha}$ respectively denote the loss over the output and the attention; $\gamma$ is a scaling factor. As an alignment corresponds to a categorical distribution, KL-divergence is a natural difference metric. By default, the two models are trained separately. $\bm{\theta}$ is trained in teacher forcing mode, and then fixed to generate the reference attention. $\hat{\bm{\theta}}$ is trained with the joint loss $\mathcal{L}_{y,\alpha}$.

\subsection{Scheduled attention forcing}\label{sec:ch_af_sched}

\begin{figure}%
    \centering
    \includegraphics[width=5cm]{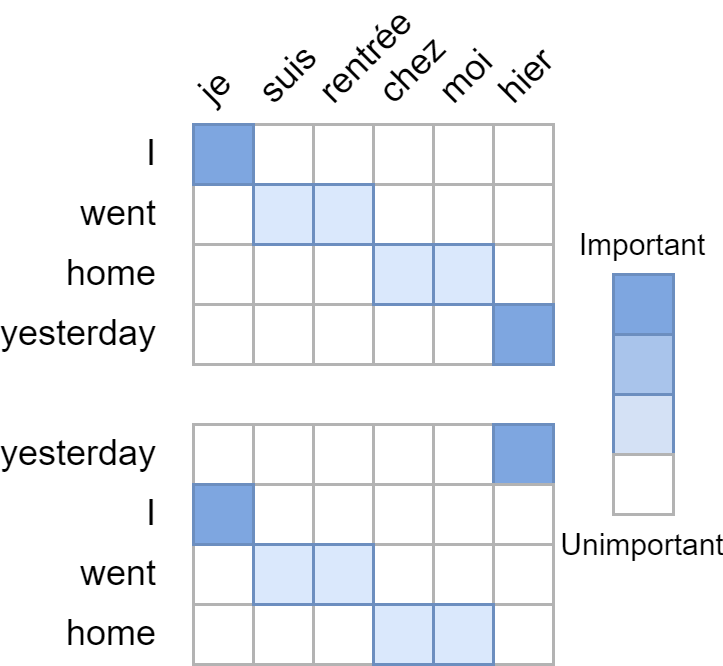}%
    \caption{Alignments: up) $\bm{\alpha}_{1:T}$ between the input and the reference output; down) $\hat{\bm{\alpha}}_{1:T}$ between the input and the generated output. For the input \emph{``je suis rentrée chez moi hier''}, the reference output is \emph{``I went home yesterday''}. When using attention forcing, the model is guided by the generated back-history and outputs \emph{``yesterday I went home''}. Here the alignment $\bm{\alpha}_{1:T}$ is not a sensible target for $\hat{\bm{\alpha}}_{1:T}$.} \label{align-TF-AF}
\end{figure}

\paragraph{Motivation}
When applying attention forcing, it is important to consider the nature of the attention connecting the input and output. For some tasks, the attention is expected to be monotonic, and there is only one type of valid attention maps, where the important positions roughly form a diagonal line. Examples include ASR and TTS. For other tasks, there may be multiple valid modes of attention: the ordering of tokens can be changed while the output sequence remains correct. Examples include NMT and text summarization. If the model takes an ordering that is different from the reference output, the token-level losses will be misleading. Figure \ref{align-TF-AF} illustrates the problem with an NMT example.

Furthermore, there might be other issues such as grave mistakes. For tasks where the output is continuous,\footnote{
The meaning of ``continuous'' comes in two folds. First, speech is continuous in time, although often sampled as a discrete sequence. For a speech sequence, the correlation between tokens is stronger than that in a text sequence. Second, a speech token follows a continuous distribution. A text token follows a discrete distribution.
}
such as TTS and voice conversion, a small deviation from the reference output is usually not a serious problem. However, this is more serious for tasks where the output is discrete, such as NMT and text summarization. During training, errors in the output history can be so serious that the token-level target is not appropriate, often due to misalignment between the generated output and the reference output. To illustrate the problem, suppose the reference output is \emph{``thank you for listening''}, and the model predicts \emph{``thanks''} at the first time step. In this case, the next output should not be \emph{``you''}, and \emph{``for''} would be a more sensible target.


\paragraph{Framework}

Scheduled attention forcing is proposed for applications where attention forcing, also referred to as ``vanilla attention forcing'', may result in an inappropriate loss. The basic idea is to automatically decide, for each input-output pair in the training data, whether vanilla attention forcing will be used. This is realized by tracking the alignment between the reference and the generated output. If they are relatively well-aligned, vanilla attention forcing will be used, otherwise a more stable training mode will be used.

\begin{figure}
\centering
\includegraphics[width=7.5cm]{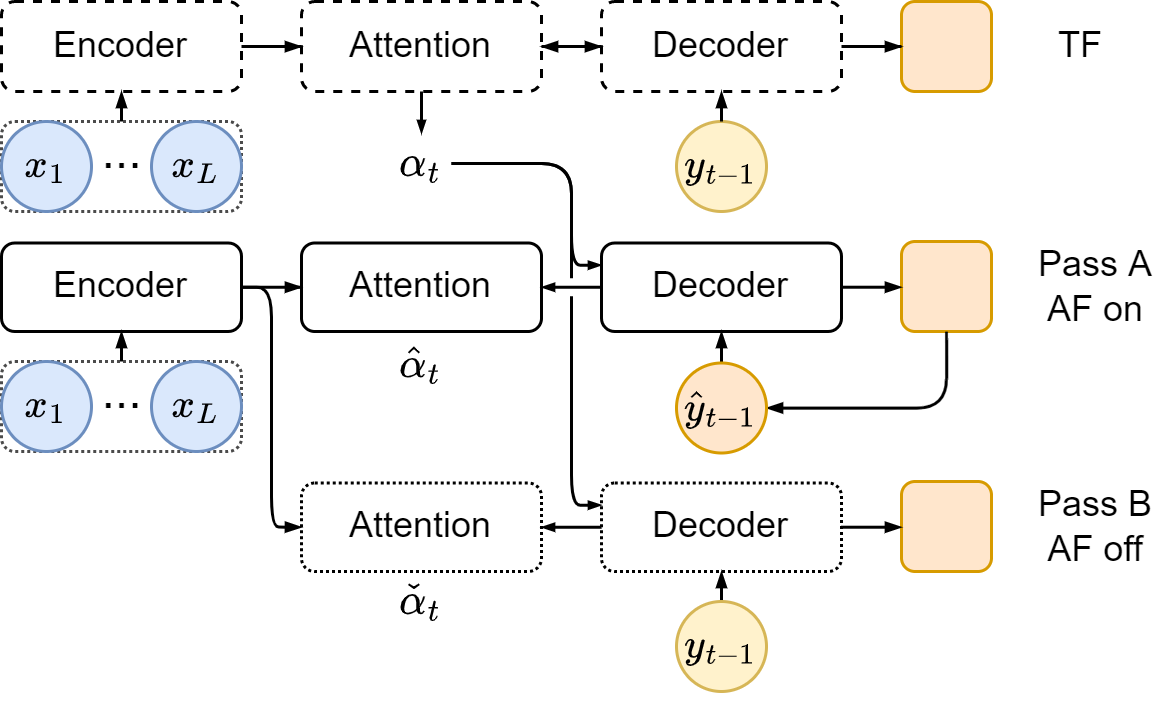}
\caption{Scheduled attention forcing. Passes A and B share the same model parameters; only one of them will be used in back-propagation, depending on the data.} \label{fig:AAF}
\end{figure}

Figure \ref{fig:AAF} illustrates scheduled attention forcing. For each input sequence, the attention forcing model $\hat{\bm{\theta}}$ takes two forward passes. Pass A is guided by the generated output history $\hat{\bm{y}}_{1:t-1}$, which is the same as vanilla attention forcing. Pass B is guided by the reference output history $\bm{y}_{1:t-1}$. The reference attention is always used, so the context vector $\hat{\bm{c}}_{t}$ is the same in both passes. If memory permits, the two forward passes can be completed in parallel, resulting in no extra time. This can be formulated as follows. For vectors produced in pass A, the notation has the hat $\hat{}$ accent; the equivalent for pass B is the check $\check{}$ accent.
\begin{gather}
\small\begin{matrix}
&\bm{h}_{1:L} = f(\bm{x}_{1:L}; \bm{\theta}_{h}) &\hat{\bm{h}}_{1:L} = f(\bm{x}_{1:L}; \hat{\bm{\theta}}_{h})
\end{matrix}\\
\small \begin{matrix}
&\bm{s}_{t} = f(\bm{y}_{1:t-1}; \bm{\theta}_{s}) &\begin{matrix}
\hat{\bm{s}}_{t} = f(\hat{\bm{y}}_{1:t-1}; \hat{\bm{\theta}}_{s})\\ 
\check{\bm{s}}_{t} = f(\bm{y}_{1:t-1}; \hat{\bm{\theta}}_{s})
\end{matrix}
\end{matrix}\\
\small \begin{matrix}
&\bm{\alpha}_{t} = f(\bm{s}_{t}, \bm{h}_{1:L}; \bm{\theta}_{\alpha}) &\begin{matrix}
\hat{\bm{\alpha}}_{t} = f(\hat{\bm{s}}_{t}, \hat{\bm{h}}_{1:L}; \hat{\bm{\theta}}_{\alpha})\\ 
\check{\bm{\alpha}}_{t} = f(\check{\bm{s}}_{t}, \hat{\bm{h}}_{1:L}; \hat{\bm{\theta}}_{\alpha})
\end{matrix}
\end{matrix}\\
\hat{\bm{c}}_{t} =\textstyle \sum_{l=1}^{L} \alpha_{t,l} \hat{\bm{h}}_{l} \\
\begin{matrix}
\hat{\bm{y}}_{t} \sim p(\cdot | \hat{\bm{s}}_{t}, \hat{\bm{c}}_{t}; \hat{\bm{\theta}}_{y})\\ 
\check{\bm{y}}_{t} \sim p(\cdot | \check{\bm{s}}_{t}, \hat{\bm{c}}_{t}; \hat{\bm{\theta}}_{y})
\end{matrix}
\end{gather}

Next, the choice of training mode is made at the sequence level. If $\textstyle \sum_{t=1}^{T} \mathrm{KL}( \bm{\alpha}_{t} || \hat{\bm{\alpha}}_{t}; \hat{\bm{\theta}}) < \lambda \sum_{t=1}^{T} \mathrm{KL}( \bm{\alpha}_{t} || \check{\bm{\alpha}}_{t}; \hat{\bm{\theta}})$, meaning that $\hat{\bm{y}}_{1:T}$ is well aligned with $\bm{y}_{1:T}$, pass A will be used in the back-propagation. The loss is the same as in vanilla attention forcing, shown in equation \ref{eq:loss_joint_AF}. Otherwise pass B will be used:
\begin{align}
\mathcal{L}_{y,\alpha}(\hat{\bm{\theta}}) &= \textstyle \sum_{t=1}^{T} \log p(\bm{y}_{t} | \bm{x}_{1:L}, \bm{y}_{1:t-1}, \bm{\alpha}_{t}; \hat{\bm{\theta}}) \nonumber \\
&+ \textstyle \gamma \sum_{t=1}^{T} \mathrm{KL}( \bm{\alpha}_{t} || \check{\bm{\alpha}}_{t}; \hat{\bm{\theta}} )
\end{align}

The KL attention loss is used to determine if the alignment is good enough between the reference output $\bm{y}_{1:T}$ and the generated output $\hat{\bm{y}}_{1:T}$. As both $\bm{\alpha}_{t}$ and $\check{\bm{\alpha}}_{t}$ are computed using $\bm{y}_{1:t-1}$, they are expected to be similar, yielding a relatively small $\mathrm{KL}( \bm{\alpha}_{t} || \check{\bm{\alpha}}_{t}; \hat{\bm{\theta}})$. In contrast, $\hat{\bm{\alpha}}_{t}$ is computed using $\hat{\bm{y}}_{1:t-1}$, and $\mathrm{KL}( \bm{\alpha}_{t} || \hat{\bm{\alpha}}_{t}; \hat{\bm{\theta}})$ is expected to be larger. $\lambda$ is a hyper-parameter controlling how much out-of-alignment $\hat{\bm{y}}_{1:T}$ and $\bm{y}_{1:T}$ can be. If $\lambda \to +\infty$, scheduled attention forcing will be the same as vanilla attention forcing.

For each pair of training data, scheduled attention forcing makes a choice whether to guide the model with the reference output history or the generated output history. This approach is named ``scheduled attention forcing'', because scheduled sampling also selectively uses the generated output history. For scheduled sampling, the selection is random. For scheduled attention forcing, the selection depends on the data.

\subsection{Parallel attention forcing} \label{sec:chAF_paf}

\paragraph{Motivation}

Transformer-style models have achieved state-of-the-art performance in various tasks including NMT and TTS. For such models with a large number of parameters, parallel training is essential \cite{vaswani2017attention}. When teacher forcing is used, there are no recurrent connections in the model, and training can be done in parallel across the length $T$ of the output $\bm{y}_{1:T}$. This is more obvious when teacher forcing is rewritten as $\hat{\bm{y}}_{t} \sim p(\cdot | \bm{y}_{1:t-1}, \bm{\alpha}_{t}, \bm{x}_{1:L}; \bm{\theta})$, where $\bm{\alpha}_{t} = f(\bm{y}_{1:t-1}, \bm{x}_{1:L}; \bm{\theta})$.
The reference output history $\bm{y}_{1:t-1}$ is available for any $t$, so $\hat{\bm{y}}_{1:T}$ can be computed in parallel. 

Attention forcing can be rewritten in a similar fashion: $\hat{\bm{y}}_{t} \sim p(\cdot | \hat{\bm{y}}_{1:t-1}, \bm{\alpha}_{t}, \bm{x}_{1:L}; \hat{\bm{\theta}})$, where $\hat{\bm{\alpha}}_{t} = f(\hat{\bm{y}}_{1:t-1}, \bm{x}_{1:L}; \hat{\bm{\theta}})$.
The model is guided with generated output history $\hat{\bm{y}}_{1:t-1}$. $\hat{\bm{y}}_{1:T}$ is not available beforehand, and is generated sequentially. So when applying attention forcing to Transformer-based models, training is no longer parallel.\footnote{
Transformer-based models have multiple cross attention mechanisms connecting the encoder and decoder. So when applied to these models, attention forcing involves a group of the reference attention $\bm{\alpha}_{t}^{(1:N,1:H)}$ and generated attention $\hat{\bm{\alpha}}_{t}^{(1:N,1:H)}$, where $N$ is the number of decoder layers, and $H$ the number of heads in each layer. These superscripts are omitted, to simplify the notation and to facilitate comparison with other forms of attention forcing.
}

\paragraph{Framework}

\begin{figure}
\centering
\includegraphics[width=7.5cm]{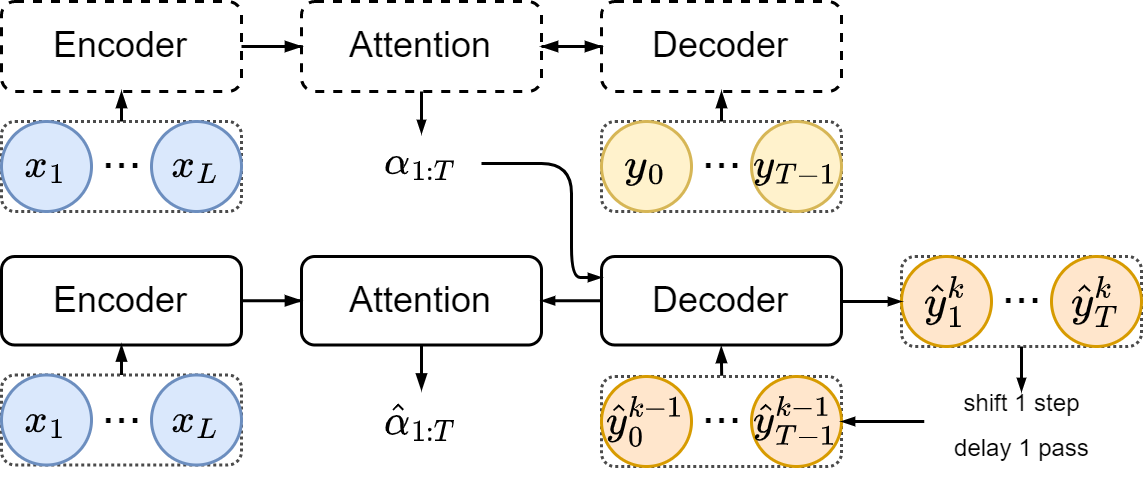}
\caption{Parallel attention forcing; at pass $k$, $\hat{\bm{y}}_{1:T}^{k-1}$ is available, so $\hat{\bm{y}}_{1:T}^{k}$ can be computed in parallel.} \label{fig:PAF}
\end{figure}

For parallel attention forcing, the core idea is to approximate the sequential generation of $\hat{\bm{y}}_{1:T}$ with a parallelizable process. Here the output $\hat{\bm{y}}_{1:T}^{K}$ is generated iteratively in $K$ forward passes. For each pass, the complete output history is available beforehand, so training can be run in parallel across time $t$, as illustrated by figure \ref{fig:PAF}.

For the first pass, the output history is the reference $\bm{y}_{1:T}$. For the following passes, the output history is the output of the previous pass $\hat{\bm{y}}_{1:T}^{k-1}$.
\begin{gather}
\hat{\bm{y}}_{1:T}^{0} = \bm{y}_{1:T} \\
\hat{\bm{\alpha}}_{t}^{k} = f(\hat{\bm{y}}_{1:t-1}^{k-1}, \bm{x}_{1:L}; \hat{\bm{\theta}}) \\
\hat{\bm{y}}_{t}^{k}
\left\{\begin{matrix}
= &\hat{\bm{y}}_{t}^{k-1} &\text{ if } t<k\\ 
\sim &p(\cdot | \hat{\bm{y}}_{1:t-1}^{k-1}, \bm{\alpha}_{t}, \bm{x}_{1:L}; \hat{\bm{\theta}}) &\text{ if } t \geq k
\end{matrix}\right.
\end{gather}
It can be proved that when $K=T$, $\hat{\bm{y}}_{1:T}^{K}$ is independent of the reference back-history, and is equivalent to an output sequentially generated \cite{duckworth2019parallel}. In appendix \ref{sec:appendix_af}, figure \ref{fig:parallel} illustrates how the iterative parallel generation approximates sequential generation. Empirically, $K$ could be much smaller than $T$, while still addressing the exposure bias \cite{duckworth2019parallel}. So although parallel attention forcing requires more computation than vanilla attention forcing, it is more efficient thanks to parallel training.

Attention forcing has a regularizing effect. During training, the attention mechanism(s) of the attention forcing model is encouraged to mimic the teacher forcing model. Hence there is the risk of over regularization, in which case the attention forcing model converges to the teacher forcing model. When applying attention forcing to Transformer-based models, our default option is to force all the cross-attention connecting the encoder and the decoder, while leaving the self-attention alone. However, in a Transformer-based model, there are usually dozens of such attention maps. The exact number is equal to the number of decoder layers times the number of attention heads in each layer. In contrast, in a model based on RNN or CNN, there is only one attention map. Forcing all the attention heads tends to over regularize Transformer-based models.
This problem can be addressed by forcing selected attention heads only. For the selected attention heads, the reference attention is given to the following layer, and an alignment loss is computed between the reference and the predicted attention. For the other attention heads, the predicted attention is given to the following layer as usual. In appendix \ref{sec:appendix_af}, figure \ref{fig:chAF_paf_mask} illustrates the idea of forcing selected attention heads. For Transformer-based models, different attention heads have different functions \cite{voita2019analyzing, vig2019analyzing}. For example, attention heads in the deepest layers of the model capture the most distant relationships \cite{vig2019analyzing}. In this paper, the selection is mainly based on the layer.

\subsection{Related work}
Attention forcing follows the first line of approaches addressing exposure bias, descried in section \ref{sec:intro_inf_train}. Similar to scheduled sampling and professor forcing, it is between teacher forcing and free running. An advantage of attention forcing is that it does not require a heuristic schedule or a discriminator, which can be difficult to tune. \cite{lamb2016professor} reported negative results on TTS. Variations of scheduled sampling were applied to NMT, resulting in both positive \cite{zhang2019bridging} and negative \cite{duckworth2019parallel} results.


Compared with sequence-level training approaches, attention forcing is more efficient, in the sense that it does not require generating multiple output sequences during training. Reference \cite{shen2016minimum} applied Minimum Bayes Risk (MBR) training to NMT, and approximates the expectation of the risk by sampling and renormalizing the probability of the samples. References \cite{ranzato2015sequence, bahdanau2016actor} approximate the same loss with Monte Carlo sampling, and optimizes the loss using Reinforcement Learning (RL). 

For sequence-level training, another general concern is the choice of the distance metric, i.e. the risk. Many tasks, including NMT and TTS, do not have a gold-standard objective metric. Empirically, models trained with one metric may not perform equally well when assessed using another metric \cite{ranzato2015sequence}. To tackle this issue, adversarial training \cite{yu2017seqgan, wu2018adversarial} can be used: a discriminator learns a loss function, which is potentially better than standard metrics. The difficulty here is that the discriminator itself can be difficult to train \cite{zhang2018bidirectional}. While attention forcing does not directly optimize a sequence-level loss, it can indirectly reduce the loss by training the model to recover from errors. This is because sequence-level metrics are usually computed by comparing units of the sequences. Examples include word error rate for ASR, and BLEU \cite{papineni2002bleu} and ROUGE \cite{lin2003automatic} for NMT. If models can recover from its errors, the sequence-level loss can be reduced.

It is challenging to apply attention forcing to models without an attention mechanism. However, the concept of attention forcing can be applied to such models, where it is essential to find something analogous to attention. For convolutional neural networks, for example, attention maps can be defined based on the activation or gradient \cite{zagoruyko2016paying}.

On a historical note, attention forcing was first proven successful in TTS \cite{dou2019attention, dou2020attention}. Then scheduled attention forcing was introduced for machine translation in our preprint paper \cite{dou2021attention}. At the time it was named "automatic attention forcing". We find it best to introduce both scheduled attention forcing and parallel attention forcing in a single paper as they are applied together for Transformer-style NMT models.

\section{Experiments} \label{sec:chNMT_exp}

\paragraph{Data} \label{sec:chNMT_exp_data}
There are two sources of data: IWSLT'15 \cite{cettolo2012wit3, cettolo2015iwslt} and WMT'16 \cite{bojar2016findings}. Table \ref{tab:chNMT_exp_data} shows the data split. Detailed descriptions are in appendix \ref{sec:appendix_exp}. The sentences in IWSLT are from TED talks. Two translation directions are investigated: English-to-French (EnFr) and English-to-Vietnamese (EnVi). The data preprocessing follows \cite{luong2015stanford}. 
The sentences in WMT are from newspaper articles. Here English-to-German (EnDe) translation is investigated. The data preprocessing follows \cite{ott2018scaling}.
For all the translation directions, the Moses tokenizer \cite{koehn2007moses} is adopted, and the translations are detokenized before evaluation. The checkpoints are selected based on the validation set, and the results are compared on the test set.
\begin{table}
\footnotesize
\centering
\caption{\label{tab:chNMT_exp_data} Data used in the experiments.}
\begin{tabular}{@{}l|l|l}
\toprule
        &             & \# sentence pairs \\
        & Languages & Training-valid-test       \\ \midrule
IWSLT'15 & En$\to$Fr   & 208K-1026-1305       \\
        & En$\to$Vi   & 133K-1553-1268       \\ \midrule
WMT'16   & En$\to$De   & 4.5M-3000-3003       \\ \bottomrule
\end{tabular}
\end{table}

\paragraph{Performance Metrics} \label{sec:chNMT_exp_eval}

The overall translation quality is measured by BLEU \cite{papineni2002bleu}. The average of 1-to-4 gram BLEU scores are computed and a 0.6 brevity penalty is applied. For IWSLT and WMT data, the BLEU score is computed using the Moses toolkit \cite{koehn2007moses} and the SacreBLEU toolkit \cite{post2018call}, respectively.

In NMT, there can be multiple valid output sequences for a single input sequence. Given that the overall translation quality is the same, it is desirable for an NMT model output to be diverse translations for the same input. This work measures the diversity of the candidate translations by pairwise BLEU \cite{shen2019mixture} and entropy. For a translation model $\bm{\theta}$, we use sampling search $M$ times with different random seeds, obtaining a group of translations $\{ \hat{\bm{y}}^{(m)} \}_{m=1}^{M}$. $\hat{\bm{y}}^{(m)}$ denotes all the output sentences in the dev or test set. Then we compute the average BLEU between all pairs: $\frac{1}{M(M-1)} \sum_{n=1}^{M} \sum_{m=1}^{M} {\tt BLEU} (\hat{\bm{y}}^{(n)}, \hat{\bm{y}}^{(m)})_{n \neq m}$. In our experiments, $M$ is set to 5. The more diverse the translations, the lower the pairwise BLEU. In addition to pairwise BLEU, we use greedy search and save the entropy $e_{t}$ of the output token's distribution at each time step. Let $e_{1:T}$ cover all the model output steps, we compute the average value: $e = \frac{1}{T} \sum_{t=1}^{T} e_{t}$. Higher entropy means that the model is less certain, and thus more likely to produce diverse outputs. This process is deterministic, and is not repeated with different random seeds.

\subsection{Scheduled attention forcing} \label{sec:chNMT_exp_saf}
\paragraph{Setup}
The experiments are conducted with EnFr and EnVi data in IWSLT'15. The model is based on GNMT \cite{wu2016google}. The details of the model and training are described in appendix \ref{sec:appendix_exp_saf}. By default, the baseline models are trained with Teacher Forcing (TF) for fewer than 60 epochs. Starting from the baseline, models are finetuned with Attention Forcing (AF) for fewer than 30 epochs. For AF, the scale $\gamma$ of the attention loss is 10. The default inference approach is greedy search. When investigating diversity, sampling search is also adopted. The checkpoints are selected based on the validation BLEU. For all the training approaches, the effective number of epochs is smaller than the maximum, i.e. training goes on until convergence. The computational budget to train a model is 96 hours on a single Nvidia Tesla P100.

\paragraph{Results}
First, we compare TF and vanilla AF. The preliminary experiments show that when pretraining with TF is adopted, the BLEU of AF increases from 21.77 to 22.93 for EnFr, and from 13.92 to 18.27 for EnVi. However, it does not outperform TF, as shown by the first two rows in each section of table \ref{tab:BLEU_TF_AAF_tune}. This result is expected, considering the discrete and multi-modal nature of the NMT output space, analyzed in section \ref{sec:ch_af_sched}.
\begin{table} \footnotesize
\centering
\caption{BLEU of Teacher Forcing (TF), Attention Forcing (AF) and Scheduled Attention Forcing (SAF) with different values of $\lambda$; the higher $\lambda$ is, the more likely the generated output history is used; the models are based on GNMT, and trained with data from IWSLT'15.}
\begin{tabular}{l|l|l|ll}
\toprule
Task  & Training  & $\lambda$     & BLEU $\uparrow$ &              \\ \midrule
EnFr  & TF  & -           & 30.70        \\
      & AF  & -           & 22.93        \\
      & SAF  & 2.5     & 31.44        \\
      & SAF  & 3.0     & \textbf{31.66}        \\
      & SAF  & 3.5     & 31.34        \\ \midrule
EnVi  & TF  & -           & 25.57        \\
      & AF  & -           & 18.27        \\
      & SAF  & 2.5   & 26.02        \\ 
      & SAF  & 3.0   & 25.71        \\ 
      & SAF  & 3.5   & \textbf{26.72}        \\ \bottomrule
\end{tabular}
\label{tab:BLEU_TF_AAF_tune}
\end{table}

Next, TF is compared with Scheduled Attention Forcing (SAF). The hyper parameter $\lambda$, introduced in section \ref{sec:ch_af_sched}, controls the tendency to use generated outputs. As shown in table \ref{tab:BLEU_TF_AAF_tune}, with very limited tuning, SAF outperforms TF. The performance is robust in a certain range of $\lambda$. In the following experiments, $\lambda$ is set to 3.0 for EnFr and 3.5 for EnVi. To reduce the randomness of the experiments, both TF and SAF are run $R=5$ times with different random seeds. Let $\{ \bm{\theta}^{(r)} \}_{r=1}^{R}$ denote the group of TF models, and $\{ \hat{\bm{\theta}}^{(r)} \}_{r=1}^{R}$ the SAF models. For both groups, the BLEU's mean $\pm$ standard deviation is computed. Table \ref{tab:BLEU_TF_AAF} shows the results. In terms of mean BLEU, SAF yields a 0.44 gain for EnFr, and a 0.55 gain for EnVi.
\begin{table}\footnotesize
\setlength{\tabcolsep}{4pt}
\centering
\caption{BLEU, Entropy and pairwise (P.) BLEU of TF and SAF; each approach is run 5 times, and the mean $\pm$ std is shown; the models are based on GNMT, and trained with data from IWSLT'15.}
\begin{tabular}{l|l|l|ll}
\toprule
Task  & Training     & BLEU$\uparrow$  & Entropy$\uparrow$ & P. BLEU$\downarrow$              \\ \midrule
EnFr  & TF           & 31.10 $\scriptstyle \pm 0.27$  & 1.060 $\scriptstyle \pm 0.047$           & 27.43 $\scriptstyle \pm 0.75$        \\
      & SAF     & \textbf{31.54} $\scriptstyle \pm 0.14$  & 1.034 $\scriptstyle \pm 0.013$           & 27.82 $\scriptstyle \pm 0.67$        \\ \midrule
EnVi  & TF           & 25.86 $\scriptstyle \pm 0.44$  & 1.508 $\scriptstyle \pm 0.012$           & 22.11 $\scriptstyle \pm 0.34$        \\
      & SAF   & \textbf{26.41} $\scriptstyle \pm 0.33$  & \textbf{1.582} $\scriptstyle \pm 0.017$              & \textbf{20.75} $\scriptstyle \pm 0.29$        \\ \bottomrule
\end{tabular}
\label{tab:BLEU_TF_AAF}
\end{table}


To measure the diversity of the translations, the entropy and pairwise BLEU are computed for $\{ \bm{\theta}^{(r)} \}_{r=1}^{R}$ and $\{ \hat{\bm{\theta}}^{(r)} \}_{r=1}^{R}$, as shown in the last two columns of table \ref{tab:BLEU_TF_AAF}. Focusing on the entropy column, SAF leads to higher entropy for EnVi, which indicates higher diversity. For EnFr, SAF and TF lead to similar levels of diversity, especially when the standard deviation is considered. We believe that the difference is due to the nature of the tasks. While English and French have similar syntax and lexicon, English and Vietnamese are more different. When trained with SAF, the EnVi model benefits more from using generated back-history, which is more likely to be different from the reference back-history. The pairwise BLEU shows similar trends. For EnVi, SAF leads to lower pairwise BLEU, i.e. higher diversity. For EnFr, the difference between SAF and TF is negligible. So in the following experiments, only pairwise BLEU will be reported.


\subsection{Parallel attention forcing} \label{sec:chNMT_exp_paf}
\paragraph{Setup}
The experiments in this section are conducted with WMT'16 EnDe data. Compared with IWSLT, WMT is more suitable for Transformer models in terms of the amount of data.
The Translation models have the same structure as the “big” Transformer in \cite{vaswani2017attention}, and the training follows \cite{ott2018scaling}. The details are in appendix \ref{sec:appendix_exp_paf}.
The baseline models are trained with Teacher Forcing (TF). Starting from the baseline, other models are finetuned respectively with sequence-level Scheduled Sampling (SS) and Attention Forcing (AF). To keep the benefit of parallel training, AF and SS are approximated by their parallel version, as described in section \ref{sec:chAF_paf} and reference \cite{duckworth2019parallel}. The number of iterations is two. For SS, the probability of using the reference output decreases linearly from 1 to 0.7; more aggressive schedules are found to degrade the performance. For AF, the scale $\gamma$ of the attention loss is 1000. The default inference approach is beam search with beam size 4. The validation BLEU is monitored to select checkpoints and to stop training when no performance gain is observed after 10 epochs. The computational budget to train a model is 144 hours on a single Nvidia Tesla V100.

\paragraph{Results}
The first two sections of table \ref{tab:chNMT_exp_paf_sched} lists the preliminary results of TF, parallel SS and parallel AF. Here all the attention heads are constantly forced, regardless of the alignment between the reference and generated output history. Compared with TF, parallel SS yields lower BLEU as well as pairwise BLEU. It is difficult to conclude whether the decrease in pairwise BLEU results from higher diversity or lower translation quality. In its parallel version, AF performs similarly to TF. This is probably because the back-history is generated in TF mode.
As analyzed in section \ref{sec:ch_af_sched}, when applying AF to NMT, it is important to turn AF on and off based on the alignment between the reference and the generated outputs. Hence unless otherwise mentioned, a schedule is added to parallel AF in the following experiments. The last section of table \ref{tab:chNMT_exp_paf_sched} lists the performance of parallel AF, where the hyperparameter $\lambda$ of the schedule is tuned. While the BLEU remains at the same level, the pairwise BLEU decreases when the percentage of AF decreases, signaling that AF regularizes the translation model to operate in a safe zone.


\begin{table}\footnotesize
\centering
\caption{\label{tab:chNMT_exp_paf_sched} BLEU and Pairwise BLEU of Teacher Forcing (TF), Parallel Scheduled Sampling (PSS) and Parallel Attention Forcing (PAF); higher $\lambda$ means higher tendency to use AF; the models Transformer-based, trained with WMT'16 EnDe, tested on newstest14; the bold numbers correspond to the $\lambda$ for further experiments.}
\begin{tabular}{lc|c|c}
\toprule
 & $\lambda$  & BLEU$\uparrow$   & Pairwise BLEU$\downarrow$  \\ \midrule
TF      & -    & 28.68   & 31.12    \\
PSS      & -    & 28.19   & 30.17    \\ \midrule
PAF      & $+\infty$    & 28.74   & 31.90    \\ \midrule
PAF      &1.1    & 28.75    & 31.60     \\
PAF      &1.2    & \textbf{28.57}   & \textbf{30.62}    \\
PAF      &1.3    & 28.48   & 31.89    \\
PAF      &1.4    & 28.56   & 31.99    \\
PAF      &1.5    & 28.47   & 32.06    \\\bottomrule
\end{tabular}
\end{table}

As analyzed in section \ref{sec:chAF_paf}, the encoder and decoder are connected by multiple attention mechanisms. It is likely that too much information is passed from the TF baseline to the AF model. To reduce this information, we only force selected attention heads. The first two decoder layers are selected, and the reason is discussed in section \ref{sec:appendix_exp_paf} in the appendix. In each layer, the number of heads forced are 8, 12 or 16 out of 16. Table \ref{tab:chNMT_exp_paf_sched_select} lists the results. When only two layers are forced, the performance of parallel AF surpasses TF in both BLEU and pairwise BLEU. The best performance (first row) is achieved when 8 heads are forced in each layer. To reduce the randomness of hyperparameter tuning, we run another experiment where the other 8 heads are forced, and the result (second row) is comparable to the best performance.


\begin{table}\footnotesize
\setlength{\tabcolsep}{4pt}
\centering
\caption{\label{tab:chNMT_exp_paf_sched_select} BLEU and Pairwise BLEU of TF, PSS and PAF, where only selected attention heads are forced; the models are based on Transformer, trained with WMT'16 EnDe, tested on newstest14.}
\begin{tabular}{lc|cc|c|c}
\toprule
 & $\lambda$  & Layers  & Heads  & BLEU$\uparrow$   & Pairwise BLEU$\downarrow$  \\ \midrule
TF      & -    & -      & -      & 28.68   & 31.12    \\
PSS      & -    & -      & -      & 28.19   & 30.17    \\ \midrule
PAF      &1.2    & 1-2      & 1-8      & \textbf{29.04}   & \textbf{30.47}    \\
PAF      &1.2    & 1-2      & 9-16      & \textbf{28.91}   & \textbf{30.05}    \\
PAF      &1.2    & 1-2      & 1-12      & 28.86   & 30.87    \\
PAF      &1.2    & 1-2      & 1-16      & 28.64   & 30.79     \\\bottomrule
\end{tabular}
\end{table}

\section{Conclusion}
This paper introduced two extensions to attention forcing, a training approach addressing exposure bias in attention-based seq2seq models. We recommend the basic form of attention forcing in tasks with continuous outputs like TTS. For tasks with discrete outputs like NMT, we recommend scheduled attention forcing. For models based on Transformers, it is essential to use parallel attention forcing, and to not force all the attention heads.





\section{Broader impacts}

The training approaches introduced in this work can be applied to a range of attention-based seq2seq models. NMT is used as a representative task, where the output is discrete. The baselines \cite{wu2016google, vaswani2017attention} in the experiments are prominent models based on RNNs or Transformers, which are widely used in various tasks \cite{tay2020efficient, dou2022improving}. While this work focuses on autoregressive models, it can also benefit non-autoregressive models. For example, autoregressive models can act as teachers in teacher-student training of non-autoregressive models. They can also generate data for semi-supervised training.

The datasets \cite{cettolo2012wit3, cettolo2015iwslt, bojar2016findings} used in the experiments are public dataset repeatedly used in the NMT community \cite{luong2015stanford, ott2018scaling}. Sentences in IWSLT are from TED talks, and sentences in WMT are from newspapers. While we did not observe any sensitive information in the data, please contact us if any potential risks are spotted in the data, such as privacy issues. The tools \cite{koehn2007moses, post2018call} used to compute BLEU scores are also public. The use of the data and tools is consistent with their intended use. Detailed documentation of the data and code is available in their references.

\bibliography{custom}

\begin{thebibliography}{35}
\expandafter\ifx\csname natexlab\endcsname\relax\def\natexlab#1{#1}\fi

\bibitem[{Bahdanau et~al.(2017)Bahdanau, Brakel, Xu, Goyal, Lowe, Pineau,
  Courville, and Bengio}]{bahdanau2016actor}
Dzmitry Bahdanau, Philemon Brakel, Kelvin Xu, Anirudh Goyal, Ryan Lowe, Joelle
  Pineau, Aaron Courville, and Yoshua Bengio. 2017.
\newblock An actor-critic algorithm for sequence prediction.
\newblock \emph{5th International Conference on Learning Representations,
  {ICLR}}.

\bibitem[{Bengio et~al.(2015)Bengio, Vinyals, Jaitly, and
  Shazeer}]{bengio2015scheduled}
Samy Bengio, Oriol Vinyals, Navdeep Jaitly, and Noam Shazeer. 2015.
\newblock Scheduled sampling for sequence prediction with recurrent neural
  networks.
\newblock In \emph{Advances in Neural Information Processing Systems}, pages
  1171--1179.

\bibitem[{Bojar et~al.(2016)Bojar, Chatterjee, Federmann, Graham, Haddow, Huck,
  Yepes, Koehn, Logacheva, Monz et~al.}]{bojar2016findings}
Ond{\v{r}}ej Bojar, Rajen Chatterjee, Christian Federmann, Yvette Graham, Barry
  Haddow, Matthias Huck, Antonio~Jimeno Yepes, Philipp Koehn, Varvara
  Logacheva, Christof Monz, et~al. 2016.
\newblock Findings of the 2016 conference on machine translation.
\newblock In \emph{Proceedings of the First Conference on Machine Translation:
  Volume 2, Shared Task Papers}, pages 131--198.

\bibitem[{Cettolo et~al.(2012)Cettolo, Girardi, and Federico}]{cettolo2012wit3}
Mauro Cettolo, Christian Girardi, and Marcello Federico. 2012.
\newblock Wit3: Web inventory of transcribed and translated talks.
\newblock In \emph{Conference of european association for machine translation},
  pages 261--268.

\bibitem[{Cettolo et~al.(2015)Cettolo, Niehues, St{\"u}ker, Bentivogli,
  Cattoni, and Federico}]{cettolo2015iwslt}
Mauro Cettolo, Jan Niehues, Sebastian St{\"u}ker, Luisa Bentivogli, Roldano
  Cattoni, and Marcello Federico. 2015.
\newblock The {IWSLT} 2015 evaluation campaign.
\newblock In \emph{Proceedings of the 12th International Workshop on Spoken
  Language Translation: Evaluation Campaign}.

\bibitem[{Chen et~al.(2018)Chen, Firat, Bapna, Johnson, Macherey, Foster,
  Jones, Schuster, Shazeer, Parmar et~al.}]{chen2018best}
Mia~Xu Chen, Orhan Firat, Ankur Bapna, Melvin Johnson, Wolfgang Macherey,
  George Foster, Llion Jones, Mike Schuster, Noam Shazeer, Niki Parmar, et~al.
  2018.
\newblock The best of both worlds: Combining recent advances in neural machine
  translation.
\newblock In \emph{Proceedings of the 56th Annual Meeting of the Association
  for Computational Linguistics (Volume 1: Long Papers)}, pages 76--86.

\bibitem[{Dou(2022)}]{dou2022improving}
Qingyun Dou. 2022.
\newblock \emph{Improving Attention-based Sequence-to-sequence Models}.
\newblock Ph.D. thesis, University of Cambridge.

\bibitem[{Dou et~al.(2020)Dou, Efiong, and Gales}]{dou2020attention}
Qingyun Dou, Joshua Efiong, and Mark~JF Gales. 2020.
\newblock Attention forcing for speech synthesis.
\newblock \emph{Proc. Interspeech 2020}, pages 4014--4018.

\bibitem[{Dou et~al.(2019)Dou, Lu, Efiong, and Gales}]{dou2019attention}
Qingyun Dou, Yiting Lu, Joshua Efiong, and Mark~JF Gales. 2019.
\newblock Attention forcing for sequence-to-sequence model training.
\newblock \emph{arXiv preprint arXiv:1909.12289}.

\bibitem[{Dou et~al.(2021)Dou, Lu, Manakul, Wu, and Gales}]{dou2021attention}
Qingyun Dou, Yiting Lu, Potsawee Manakul, Xixin Wu, and Mark J.~F. Gales. 2021.
\newblock Attention forcing for machine translation.
\newblock \emph{arXiv preprint arXiv:2104.01264}.

\bibitem[{Duckworth et~al.(2019)Duckworth, Neelakantan, Goodrich, Kaiser, and
  Bengio}]{duckworth2019parallel}
Daniel Duckworth, Arvind Neelakantan, Ben Goodrich, Lukasz Kaiser, and Samy
  Bengio. 2019.
\newblock Parallel scheduled sampling.
\newblock \emph{arXiv preprint arXiv:1906.04331}.

\bibitem[{Guo et~al.(2019)Guo, Soong, He, and Xie}]{guo2019new}
Haohan Guo, Frank~K Soong, Lei He, and Lei Xie. 2019.
\newblock A new {GAN}-based end-to-end {TTS} training algorithm.
\newblock \emph{Interspeech}.

\bibitem[{Koehn et~al.(2007)Koehn, Hoang, Birch, Callison-Burch, Federico,
  Bertoldi, Cowan, Shen, Moran, Zens et~al.}]{koehn2007moses}
Philipp Koehn, Hieu Hoang, Alexandra Birch, Chris Callison-Burch, Marcello
  Federico, Nicola Bertoldi, Brooke Cowan, Wade Shen, Christine Moran, Richard
  Zens, et~al. 2007.
\newblock Moses: Open source toolkit for statistical machine translation.
\newblock In \emph{Proceedings of the 45th annual meeting of the association
  for computational linguistics companion volume proceedings of the demo and
  poster sessions}, pages 177--180.

\bibitem[{Lamb et~al.(2016)Lamb, Goyal, Zhang, Zhang, Courville, and
  Bengio}]{lamb2016professor}
Alex~M Lamb, Anirudh Goyal Alias~Parth Goyal, Ying Zhang, Saizheng Zhang,
  Aaron~C Courville, and Yoshua Bengio. 2016.
\newblock Professor forcing: A new algorithm for training recurrent networks.
\newblock In \emph{Advances In Neural Information Processing Systems}, pages
  4601--4609.

\bibitem[{Lewis et~al.(2020)Lewis, Liu, Goyal, Ghazvininejad, Mohamed, Levy,
  Stoyanov, and Zettlemoyer}]{lewis2020bart}
Mike Lewis, Yinhan Liu, Naman Goyal, Marjan Ghazvininejad, Abdelrahman Mohamed,
  Omer Levy, Veselin Stoyanov, and Luke Zettlemoyer. 2020.
\newblock {BART}: Denoising sequence-to-sequence pre-training for natural
  language generation, translation, and comprehension.
\newblock In \emph{Proceedings of the 58th Annual Meeting of the Association
  for Computational Linguistics}, pages 7871--7880.

\bibitem[{Lin and Hovy(2003)}]{lin2003automatic}
Chin-Yew Lin and Eduard Hovy. 2003.
\newblock Automatic evaluation of summaries using n-gram co-occurrence
  statistics.
\newblock In \emph{Proceedings of the 2003 Human Language Technology Conference
  of the North American Chapter of the Association for Computational
  Linguistics}, pages 150--157.

\bibitem[{Luong et~al.(2015{\natexlab{a}})Luong, Manning
  et~al.}]{luong2015stanford}
Minh-Thang Luong, Christopher~D Manning, et~al. 2015{\natexlab{a}}.
\newblock Stanford neural machine translation systems for spoken language
  domains.
\newblock In \emph{Proceedings of the international workshop on spoken language
  translation}.

\bibitem[{Luong et~al.(2015{\natexlab{b}})Luong, Pham, and
  Manning}]{luong2015effective}
Minh-Thang Luong, Hieu Pham, and Christopher~D Manning. 2015{\natexlab{b}}.
\newblock Effective approaches to attention-based neural machine translation.
\newblock In \emph{Proceedings of the 2015 Conference on Empirical Methods in
  Natural Language Processing}.

\bibitem[{Ott et~al.(2018)Ott, Edunov, Grangier, and Auli}]{ott2018scaling}
Myle Ott, Sergey Edunov, David Grangier, and Michael Auli. 2018.
\newblock Scaling neural machine translation.
\newblock In \emph{Proceedings of the Third Conference on Machine Translation:
  Research Papers}, pages 1--9.

\bibitem[{Papineni et~al.(2002)Papineni, Roukos, Ward, and
  Zhu}]{papineni2002bleu}
Kishore Papineni, Salim Roukos, Todd Ward, and Wei-Jing Zhu. 2002.
\newblock {BLEU}: a method for automatic evaluation of machine translation.
\newblock In \emph{Proceedings of the 40th annual meeting on association for
  computational linguistics}, pages 311--318. Association for Computational
  Linguistics.

\bibitem[{Pereyra et~al.(2017)Pereyra, Tucker, Chorowski, Kaiser, and
  Hinton}]{Pereyra2017Regularizing}
Gabriel Pereyra, George Tucker, Jan Chorowski, Lukasz Kaiser, and Geoffrey~E.
  Hinton. 2017.
\newblock Regularizing neural networks by penalizing confident output
  distributions.
\newblock \emph{{CoRR}}, abs/1701.06548.

\bibitem[{Post(2018)}]{post2018call}
Matt Post. 2018.
\newblock A call for clarity in reporting {BLEU} scores.
\newblock In \emph{Proceedings of the Third Conference on Machine Translation:
  Research Papers}, pages 186--191.

\bibitem[{Ranzato et~al.(2016)Ranzato, Chopra, Auli, and
  Zaremba}]{ranzato2015sequence}
Marc'Aurelio Ranzato, Sumit Chopra, Michael Auli, and Wojciech Zaremba. 2016.
\newblock Sequence level training with recurrent neural networks.
\newblock In \emph{4th International Conference on Learning Representations,
  {ICLR}}.

\bibitem[{Shen et~al.(2016)Shen, Cheng, He, He, Wu, Sun, and
  Liu}]{shen2016minimum}
Shiqi Shen, Yong Cheng, Zhongjun He, Wei He, Hua Wu, Maosong Sun, and Yang Liu.
  2016.
\newblock Minimum risk training for neural machine translation.
\newblock In \emph{Proceedings of the 54th Annual Meeting of the Association
  for Computational Linguistics (Volume 1: Long Papers)}, pages 1683--1692.

\bibitem[{Shen et~al.(2019)Shen, Ott, Auli, and Ranzato}]{shen2019mixture}
Tianxiao Shen, Myle Ott, Michael Auli, and Marc'Aurelio Ranzato. 2019.
\newblock Mixture models for diverse machine translation: Tricks of the trade.
\newblock In \emph{ICML}.

\bibitem[{Tay et~al.(2020)Tay, Dehghani, Bahri, and Metzler}]{tay2020efficient}
Yi~Tay, Mostafa Dehghani, Dara Bahri, and Donald Metzler. 2020.
\newblock Efficient {T}ransformers: A survey.
\newblock \emph{arXiv preprint arXiv:2009.06732}.

\bibitem[{Vaswani et~al.(2017)Vaswani, Shazeer, Parmar, Uszkoreit, Jones,
  Gomez, Kaiser, and Polosukhin}]{vaswani2017attention}
Ashish Vaswani, Noam Shazeer, Niki Parmar, Jakob Uszkoreit, Llion Jones,
  Aidan~N Gomez, Lukasz Kaiser, and Illia Polosukhin. 2017.
\newblock Attention is all you need.
\newblock \emph{Advances in Neural Information Processing Systems}.

\bibitem[{Vig and Belinkov(2019)}]{vig2019analyzing}
Jesse Vig and Yonatan Belinkov. 2019.
\newblock Analyzing the structure of attention in a {T}ransformer language
  model.
\newblock In \emph{Proceedings of the 2019 ACL Workshop BlackboxNLP: Analyzing
  and Interpreting Neural Networks for NLP}, pages 63--76.

\bibitem[{Voita et~al.(2019)Voita, Talbot, Moiseev, Sennrich, and
  Titov}]{voita2019analyzing}
Elena Voita, David Talbot, Fedor Moiseev, Rico Sennrich, and Ivan Titov. 2019.
\newblock Analyzing multi-head self-attention: Specialized heads do the heavy
  lifting, the rest can be pruned.
\newblock In \emph{Proceedings of the 57th Annual Meeting of the Association
  for Computational Linguistics}, pages 5797--5808.

\bibitem[{Wu et~al.(2018)Wu, Xia, Tian, Zhao, Qin, Lai, and
  Liu}]{wu2018adversarial}
Lijun Wu, Yingce Xia, Fei Tian, Li~Zhao, Tao Qin, Jianhuang Lai, and Tie-Yan
  Liu. 2018.
\newblock Adversarial neural machine translation.
\newblock In \emph{Asian Conference on Machine Learning}, pages 534--549. PMLR.

\bibitem[{Wu et~al.(2016)Wu, Schuster, Chen, Le, Norouzi, Macherey, Krikun,
  Cao, Gao, Macherey et~al.}]{wu2016google}
Yonghui Wu, Mike Schuster, Zhifeng Chen, Quoc~V Le, Mohammad Norouzi, Wolfgang
  Macherey, Maxim Krikun, Yuan Cao, Qin Gao, Klaus Macherey, et~al. 2016.
\newblock Google's neural machine translation system: Bridging the gap between
  human and machine translation.
\newblock \emph{arXiv preprint arXiv:1609.08144}.

\bibitem[{Yu et~al.(2017)Yu, Zhang, Wang, and Yu}]{yu2017seqgan}
Lantao Yu, Weinan Zhang, Jun Wang, and Yong Yu. 2017.
\newblock Seq{GAN}: sequence generative adversarial nets with policy gradient.
\newblock In \emph{Proceedings of the Thirty-First AAAI Conference on
  Artificial Intelligence}, pages 2852--2858.

\bibitem[{Zagoruyko and Komodakis(2017)}]{zagoruyko2016paying}
Sergey Zagoruyko and Nikos Komodakis. 2017.
\newblock Paying more attention to attention: Improving the performance of
  convolutional neural networks via attention transfer.
\newblock \emph{5th International Conference on Learning Representations,
  {ICLR}}.

\bibitem[{Zhang et~al.(2019)Zhang, Feng, Meng, You, and
  Liu}]{zhang2019bridging}
Wen Zhang, Yang Feng, Fandong Meng, Di~You, and Qun Liu. 2019.
\newblock Bridging the gap between training and inference for neural machine
  translation.
\newblock In \emph{Proceedings of the 57th Annual Meeting of the Association
  for Computational Linguistics}, pages 4334--4343.

\bibitem[{Zhang et~al.(2018)Zhang, Liu, Li, Zhou, and
  Chen}]{zhang2018bidirectional}
Zhirui Zhang, Shujie Liu, Mu~Li, Ming Zhou, and Enhong Chen. 2018.
\newblock Bidirectional generative adversarial networks for neural machine
  translation.
\newblock In \emph{Proceedings of the 22nd conference on computational natural
  language learning}, pages 190--199.

\end{thebibliography}
\bibliographystyle{acl_natbib}

\appendix

\section{Details of parallel attention forcing}
\label{sec:appendix_af}

\begin{figure}
\centering
\includegraphics[width=8cm]{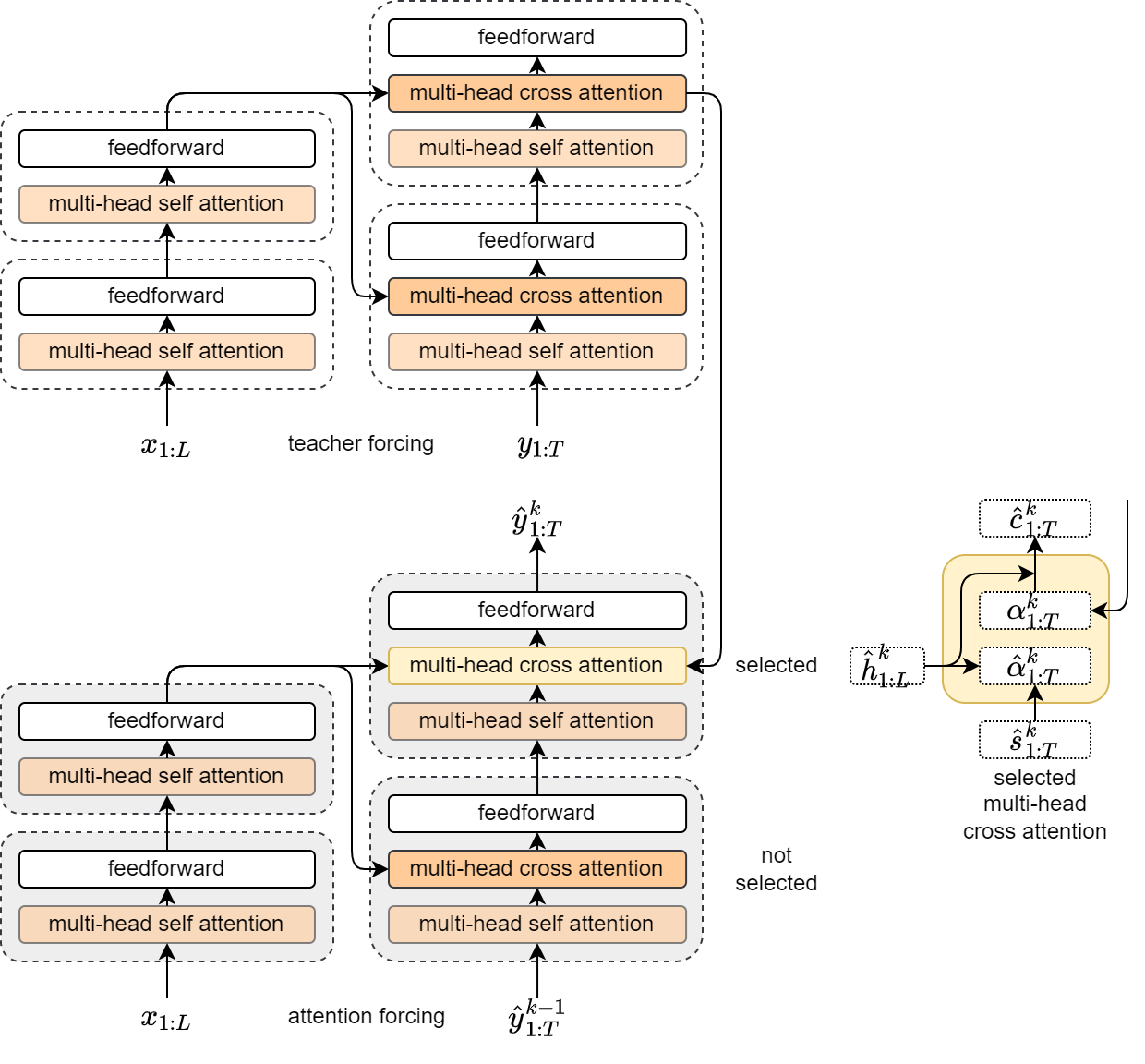}
\caption{Illustration of parallel attention forcing, applied to a Transformer with two encoder layers and two decoder layers; the attention heads in the second decoder layer are forced.}\label{fig:chAF_paf_mask}
\end{figure}

\begin{figure}
\centering
\includegraphics[width=7.5cm]{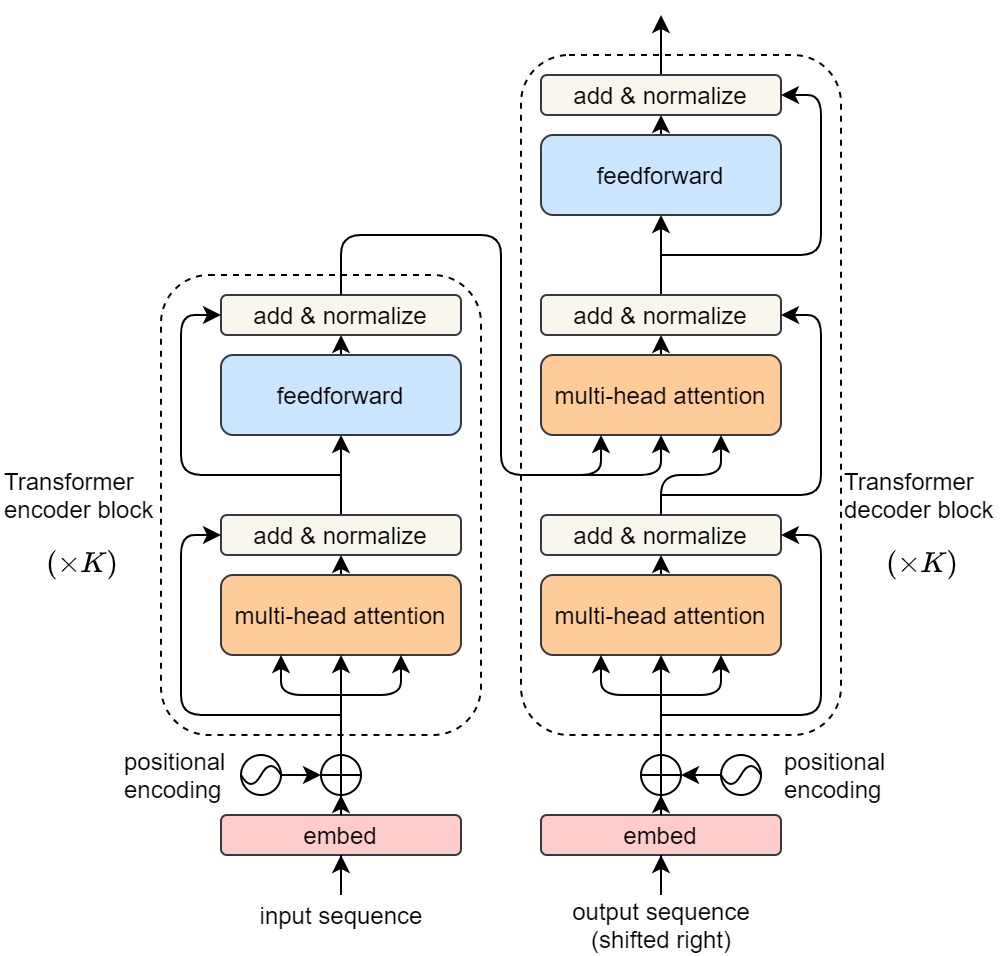}
\caption{Illustration of Transformer blocks \cite{vaswani2017attention}; the dashed rectangle in the left is a Transformer encoder block; the dashed rectangle in the right is a Transformer decoder block. }\label{fig:ch2_transformer_block}
\end{figure}

Section \ref{sec:chNMT_exp_paf} introduced parallel attention forcing, and explained the reasons to not force all the attention heads. Figure \ref{fig:chAF_paf_mask} illustrates the idea of forcing selected attention heads. Parallel attention forcing is applied to a Transformer with two decoder layers, but only forcing the attention heads in the second layer. Here the Transformer model is simplified; the detailed model structure is shown in figure \ref{fig:ch2_transformer_block}.

A Transformer block \cite{vaswani2017attention}, also referred to as a Transformer layer, is a combination of many basic modules. There are two types of Transformer blocks: Transformer encoder blocks, which encode a sequence, and Transformer decoder blocks, which connect two sequences. Figure \ref{fig:ch2_transformer_block} illustrates both types of Transformer blocks, as well as how they are combined to form an encoder-decoder model.

Similar to scheduled attention forcing, multiple forward passes can be taken for each pair of training data in parallel attention forcing. The loss function is selected based on the alignment between the reference and generated output sequences. If $ \sum_{t=1}^{T} \mathrm{KL} (\bm{\alpha}_{t} || \hat{\bm{\alpha}}_{t}^{K}; \hat{\bm{\theta}} ) < \lambda \sum_{t=1}^{T} \mathrm{KL} (\bm{\alpha}_{t} || \hat{\bm{\alpha}}_{t}^{1}; \hat{\bm{\theta}} )$, it will be assumed that $\hat{\bm{y}}_{1:T}^{K}$ is well aligned with $\bm{y}_{1:T}$, and the $K$-th forward pass will be used in the back-propagation:
\begin{align}
\mathcal{L}_{y,\alpha} (\hat{\bm{\theta}}) &=  -\sum_{t=1}^{T} \log p(\bm{y}_{t} | \hat{\bm{y}}_{1:t-1}^{K}, \bm{\alpha}_{t}, \bm{x}_{1:L}; \hat{\bm{\theta}}) \nonumber \\
&+ \gamma \sum_{t=1}^{T} \mathrm{KL} (\bm{\alpha}_{t} || \hat{\bm{\alpha}}_{t}^{K}; \hat{\bm{\theta}} )
\end{align}

Otherwise the first pass will be used:
\begin{align}
\mathcal{L}_{y,\alpha} (\hat{\bm{\theta}}) &=  -\sum_{t=1}^{T} \log p(\bm{y}_{t} | \hat{\bm{y}}_{1:t-1}^{1}, \bm{\alpha}_{t}, \bm{x}_{1:L}; \hat{\bm{\theta}}) \nonumber \\
&+\gamma \sum_{t=1}^{T} \mathrm{KL} (\bm{\alpha}_{t} || \hat{\bm{\alpha}}_{t}^{1}; \hat{\bm{\theta}} )
\end{align}

\begin{figure}
\centering
\includegraphics[width=6cm]{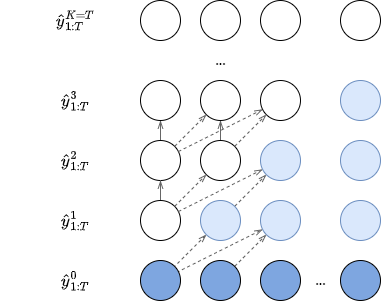}
\caption{Illustration of iterative parallel generation; the dark blue circles are the reference tokens, and the rest are generated tokens; the light blue circles are influenced by the reference, and the white circles are not; the solid arrows represent copying, the dashed arrows represent dependency. \label{fig:parallel}}
\end{figure}

Figure \ref{fig:parallel} illustrates how the iterative parallel generation approximates sequential generation. It can be proved that when $K=T$, $\hat{\bm{y}}_{1:T}^{K}$ is independent of the reference back-history, and is equivalent to an output sequentially generated \cite{duckworth2019parallel}. Empirically, $K$ could be much smaller than $T$, while still addressing the exposure bias \cite{duckworth2019parallel}. So although parallel attention forcing requires more computation than vanilla attention forcing, it is more efficient thanks to parallel training.

\section{Details of experimental setup} \label{sec:appendix_exp}

\subsection{Data}
As shown in table \ref{tab:chNMT_exp_data}, there are two sources of data: IWSLT'15 \cite{cettolo2012wit3, cettolo2015iwslt} and WMT'16 \cite{bojar2016findings}. One reference output is provided for each input. The IWSLT datasets correspond to subtitle translation tasks, where the sentences are from TED talks. Two translation directions are investigated: English-to-French (EnFr) and English-to-Vietnamese (EnVi). These datasets are relatively small, and are used to train RNN-based models. For EnFr, the training set contains 208K sentence pairs. The validation set (tst2013) and test set (tst2014) respectively contain 1026 and 1305 sentence pairs. For EnVi, the training set contains 133K sentence pairs. The validation set (tst2012) and test set (tst2013) respectively contain 1553 and 1268 sentence pairs. The data preprocessing follows \cite{luong2015stanford}. The vocabularies are at the word-level, i.e. the units are words. For EnFr, both English and French vocabularies are limited to 50K. For EnVi, the vocabulary sizes are 17K and 7.7K for English and Vietnamese.

The WMT datasets correspond to news translation tasks, where the sentences are from newspaper articles. Here English-to-German (EnDe) translation is investigated. The dataset is considerably bigger, and is used to train Transformer-based models. The training set contains 4.5M sentence pairs. The validation set (newstest13) and test set (newstest14) respectively contain 3000 and 3003 sentence pairs. The data preprocessing follows reference \cite{ott2018scaling}. A joint source and target sub-word vocabulary is built using byte pair encoding. The vocabulary is 32K BPE tokens. For all the translation directions, the Moses tokenizer \cite{koehn2007moses} is adopted, and the translations are detokenized before evaluation. The checkpoints are selected based on the validation set, and the results are compared on the test set.

\subsection{Scheduled attention forcing} \label{sec:appendix_exp_saf}

\subsubsection{Setup}
The experiments in section \ref{sec:chNMT_exp_saf} are conducted with EnFr and EnVi data in IWSLT'15. The model is based on GNMT \cite{wu2016google}. The differences are as follows. The model is simplified with a smaller number of LSTM layers due to the small scale of data: the encoder has 2 layers of bidirectional LSTM and the decoder has 4 layers of unidirectional LSTM; the attention mechanism is the general form of dot-product attention \cite{luong2015effective}; both English and Vietnamese word embeddings have 200 dimensions and are randomly initialized. Table \ref{tab:chNMT_exp_gnmt} summarizes the hyperparameters. 

\begin{table}\footnotesize
\setlength{\tabcolsep}{4pt}
\centering
\caption{\label{tab:chNMT_exp_gnmt} Hyperparameters of the RNN-based translation model \cite{wu2016google}.}
\begin{tabular}{@{}l|l@{}}
\toprule
Word embedding & 200D                                      \\ \midrule
Encoder        & 2-layer bidirectional LSTM (200D)    \\ \midrule
Attention      & General dot-product \cite{luong2015effective}             \\ \midrule
Decoder        & 4-layer unibidirectional LSTM (200D) \\ \bottomrule
\end{tabular}
\end{table}

The Adam optimiser is used with a learning rate of 0.002; $\beta_{1}=0.9, \beta_{2}=0.999, \epsilon=10^{-8}$. The maximum gradient norm is set to be 1. If there is a finetuning phase, the learning rate will be halved. The batch size is 50. Dropout is used with a probability of 0.2. By default, the baseline models are trained with Teacher Forcing (TF) for fewer than 60 epochs. Starting from the baseline, models are finetuned with Attention Forcing (AF) for fewer than 30 epochs. For AF, the scale $\gamma$ of the attention loss is 10. The default inference approach is greedy search. When investigating diversity, sampling search is also adopted, which replaces the argmax operation by sampling. The checkpoints are selected based on the validation BLEU. For all the training approaches, the effective number of epochs is smaller than the maximum, i.e. training goes on until convergence. The computational budget to train a model is 96 hours on a single Nvidia Tesla P100.

\subsection{Parallel attention forcing} \label{sec:appendix_exp_paf}

\subsubsection{Setup}
The experiments in section \ref{sec:chNMT_exp_paf} are conducted with WMT'16 EnDe data. Compared with IWSLT, WMT is more suitable for Transformer models in terms of the amount of data.
The Translation models have the same structure as the “big” Transformer in \cite{vaswani2017attention}. Table \ref{tab:chNMT_exp_transformer} shows the hyperparameters. The models are optimized with Adam using $\beta_{1}= 0.9$, $\beta_{2}= 0.98$, and $\epsilon=1e^{-8}$. Following reference \cite{ott2018scaling}, large batches are built to have a maximum of 3584 tokens. The learning rate increases linearly for 4,000 steps to $5e^{-4}$, after which it is decayed proportionally to the inverse square root of the number of steps. Label smoothing \cite{Pereyra2017Regularizing} is applied with 0.1 weight for the uniform prior distribution over the vocabulary. Dropout is applied with probability 0.3 after each attention or feedforward module. Half precision optimization techniques \cite{ott2018scaling}, are adopted to speed up training.

The baseline models are trained with Teacher Forcing (TF). Starting from the baseline, other models are finetuned respectively with sequence-level Scheduled Sampling (SS) and Attention Forcing (AF). To keep the benefit of parallel training, AF and SS are approximated by their parallel version, as described in section \ref{sec:chAF_paf} and reference \cite{duckworth2019parallel}. The number of iterations is two. For SS, the probability of using the reference output decreases linearly from 1 to 0.7; more aggressive schedules are found to degrade the performance. For AF, the scale $\gamma$ of the attention loss is 1000. The default inference approach is beam search with beam size 4. The validation BLEU is monitored to select checkpoints and to stop training when no performance gain is observed after 10 epochs. The computational budget to train a model is 144 hours on a single Nvidia Tesla V100.

\begin{table}\footnotesize
\setlength{\tabcolsep}{4pt}
\centering
\caption{\label{tab:chNMT_exp_transformer} Hyperparameters of the Transformer-based translation model \cite{vaswani2017attention}; ``FC'' stands for ``fully connected''.}
\begin{tabular}{@{}l|l@{}}
\toprule
Sub-word embedding   & 1024D                                                                                                                             \\ \midrule
Encoder              & \begin{tabular}[c]{@{}l@{}}(Multi-head self-attention → \\ Feedforward) $\times$ 6\end{tabular}                                   \\ \midrule
Decoder              & \begin{tabular}[c]{@{}l@{}}(Multi-head self-attention → \\ Multi-head cross attention → \\ Feedforward) $\times$ 6\end{tabular}   \\ \midrule
Multi-head attention & \begin{tabular}[c]{@{}l@{}}16 heads $\times$ (64D query / key / value)\\ → 1024D output \\Scaled dot-product attention\\ \cite{vaswani2017attention} \end{tabular} \\ \midrule
Feedforward          & FC-4096-ReLU → FC-1024-Linear                                                                                                     \\ \bottomrule
\end{tabular}
\end{table}

\subsubsection{Ablation studies}

\begin{table}\footnotesize
\setlength{\tabcolsep}{4pt}
\centering
\caption{\label{tab:chNMT_exp_paf_select} Ablation study on forcing selected attention heads; BLEU and Pairwise BLEU of Teacher Forcing (TF), Parallel Scheduled Sampling (PSS) and Parallel Attention Forcing (PAF) without a schedule; the models are based on Transformer, trained with WMT'16 EnDe, tested on newstest14.}
\begin{tabular}{lc|cc|c|c}
\toprule
      & $\lambda$    & Layers  & Heads    & BLEU$\uparrow$   & Pairwise BLEU$\downarrow$  \\ \midrule
TF      & -    & -      & -  & 28.68   & 31.12    \\
PSS      & -    & -      & -  & 28.19   & 30.17    \\ \midrule
PAF      & $+\infty$    & 1-2      & 1-4  & 28.38   & 31.43    \\
PAF      & $+\infty$    & 1-2      & 1-8  & 28.53   & 31.28    \\
PAF      & $+\infty$    & 1-2      & 1-12  & 28.80   & 31.85    \\
PAF      & $+\infty$    & 1-2      & 1-16  & 28.74   & 31.31    \\
PAF      & $+\infty$    & 3-4      & 1-16   & 27.94   & 31.61    \\
PAF      & $+\infty$    & 5-6      & 1-16  & 27.46   & 32.29    \\ \bottomrule
\end{tabular}
\end{table}

Table \ref{tab:chNMT_exp_paf_sched} has shown that adding a schedule itself is not enough for parallel AF to surpass TF. Another series of experiments show that limiting the information passed from the TF baseline is also not enough. In other words, the two techniques must be combined. Table \ref{tab:chNMT_exp_paf_select} shows the results of forcing selected heads, without using a schedule. As analyzed in section \ref{sec:chAF_paf}, different layers in a Transformer model perform different roles. The last three rows show that forcing layers 1 and 2 yields the best performance. The first four rows show that once the layers are selected, forcing more than four heads generally leads to better performance in BLEU and pairwise BLEU. This is the motivation behind the setup of the experiments described in section \ref{sec:chNMT_exp_paf}.

\end{document}